\documentclass[letterpaper, 10 pt, conference]{ieeeconf}  

\IEEEoverridecommandlockouts                              

\usepackage{amsmath} 
\usepackage{amssymb}  

\usepackage{booktabs}
\usepackage{graphicx}

\usepackage{algorithm}
\usepackage{algorithmicx}
\usepackage{algpseudocode}
\usepackage{tabularx}

\usepackage[capitalize]{cleveref}
\crefname{section}{Sec.}{Secs.}
\Crefname{section}{Section}{Sections}
\Crefname{table}{Table}{Tables}
\crefname{table}{Tab.}{Tabs.}

\title{\LARGE \bf
Enhancing Skin Lesion Classification Generalization with Active Domain Adaptation
}

\author{Jun Ye \\ 
    Carnegie Mellon University \\
\thanks{*This work was not supported by any organization}
}

\begin{document}

\maketitle
\thispagestyle{empty}
\pagestyle{empty}

\begin{abstract}

We propose a method to improve the generalization of skin lesion classification models by combining self-supervised learning (SSL) and active domain adaptation (ADA). The main steps of the approach include selection of an SSL pre-trained model on natural image datasets, subsequent SSL retraining on all available skin-lesion datasets, fine-tuning of the model on source domain data with labels, and application of ADA methods on target domain data. The efficacy of the proposed approach is assessed in ten skin lesion datasets with five different ADA methods, demonstrating its potential to improve generalization in settings with different amounts of domain shifts. 

{\textbf{\textit{Clinical Relevance}}}\textemdash This approach is promising in facilitating the widespread clinical adoption of deep learning models for skin lesion classification, as well as other medical imaging applications.

\end{abstract}

\section{INTRODUCTION}

The past decade has witnessed a surge in the application of artificial intelligence (AI) to skin cancer detection. Given the increasing incidence of skin cancer within an aging population and the scarcity of dermatology experts, a compelling demand exists for automated AI solutions \cite{tschandl2019comparison}. Globally, it’s estimated that around three billion people lack adequate access to skin disease medical care \cite{coustasse2019use}. Using AI can greatly alleviate the problem. Dermoscopy images are normally used by dermatologists to detect skin cancers. Consequently, computer vision (CV) methods have been applied for skin cancer classification and segmentation \cite{shetty2022skin, khan2019multimodel, khan2021attributes}. Since the International Skin Imaging Collaboration (ISIC) began organizing annual challenges in 2016, steady progress has been made in algorithm performance. Deep learning models now exceed human expert performance on the provided datasets \cite{tschandl2019comparison, combalia2022validation}. 

Despite these advances in medical imaging applications, model generalization remains a crucial hurdle for broad clinical adoption \cite{fogelberg2023domain, zech2018variable, combalia2022validation, han2022degradation, wu2022skin}. For skin cancer detection specifically, Barros et al. and Daneshjou et al. find that models perform worse on dark skin tones than light skin tones, and worse on rare diseases \cite{barros2023assessing, daneshjou2022disparities}. Han et al. observe that the dermatologist-level model’s performance significantly decreases when testing on out-of-distribution data \cite{han2022degradation}. The differences between skin lesion datasets are pervasive, stemming from various sources such as demographics, skin lesion locations, institutions, acquisition devices, and lighting conditions. Transfer learning (TL) is a technique to transfer a model’s knowledge from one domain to another. Due to the lack of large labeled datasets, a common approach is to initially train models on natural image datasets, such as ImageNet \cite{deng2009imagenet, morid2020scoping, mustafa2021supervised}. Subsequently, the model is fine-tuned on medical image datasets. There are various factors affecting a model’s transferability, including transfer techniques, training data size, domains’ similarity, and model architecture \cite{matsoukas2022what}. Here we focus on the transfer techniques, since some of the factors are out of our control in actual use cases, e.g. we have to work with a given trained model on a given dataset. 

One transfer learning technique is using self-supervised learning (SSL). Supervised learning (SL) encourages models to learn task-specific features, potentially hindering their ability to generalize to unseen data. SSL, on the other hand, enables models to learn general features that are more robust across diverse datasets \cite{azizi2022robust}. Azizi et al. propose a robust and data-efficient generalization framework for medical imaging applications by doing SSL training on in-domain medical data prior to finetuning. This approach outperforms the SL models on both in-domain and out-of-domain data. Kang et al. demonstrate similar results for four different SSL methods \cite{kang2023benchmarking}. Haggerty et al. confirm the efficacy of this approach on skin lesion datasets, demonstrating that SSL retraining on skin lesion data enhances the performance of a model pre-trained on ImageNet data with SL or SSL methods \cite{haggerty2024selfsupervised}. Another TL approach is domain adaptation (DA). DA methods aim to align the source domain (data on which the model is trained) and the target domain (data the model hasn’t encountered). Few studies have explored the application of unsupervised domain adaptation (UDA) methods for skin lesion classification \cite{wang2024achieving, chamarthi2024mitigating, gilani2024adversarial}. 

In this study, we aim to combine the SSL and DA for better transfer learning. Limited research exists on combining the SSL and DA methods. Zhao et al. combine SSL and AL without domain adaptation for skin lesion segmentation application \cite{zhao2022self}. Xu et al. combines SSL and UDA for object detection and segmentation \cite{xu2019selfsupervised}. To the best of our knowledge, this work represents the first attempt to combine SSL and DA for skin lesion classification. More specifically, we use the state-of-the-art (SOTA) SSL method, DINO, which is different from the Barlow Twins SSL method used in \cite{azizi2022robust}. Instead of using the UDA method, we apply active domain adaptation (ADA) \cite{mahapatra2024alfredo} methods. This is motivated by the actual clinical setting where the model can get feedback from human experts to iteratively improve performance \cite{useini2024automatized}. We demonstrate the effectiveness of the approach in the skin lesion classification application. This approach offers seamless integration into the clinical workflows, enabling iterative performance improvement by incorporating feedback from human experts.     

Our main contributions include:
\begin{itemize}
\item We propose a workflow to improve the generalization of the skin lesion classification model across domains by combining SSL and ADA methods.    
\item We show that SSL can be an effective UDA method and applying ADA methods can add further improvements. 
\item We demonstrate our approach’s efficacy on ten skin lesion target domains’ data with five different ADA methods.
\end{itemize}

\section{Related work}
\label{sec:related}

\subsection{Self-supervised learning}
SSL is a technique to train models without labels. It can be used to create foundational models which can be fine-tuned for different downstream tasks. The success of SSL is initially observed in natural language processing (NLP) \cite{devlin2018bert, brown2020language}, and later extended to CV \cite{doersch2015unsupervised, doersch2017multi, gidaris2018unsupervised, larsson2017colorization, pathak2016context}. Recently there has been a rapid proliferation of SSL applications in the medical imaging field \cite{azizi2021big, li2021sslp, sowrirajan2021moco, srinidhi2021improving, wolf2023selfsupervised, huang2023self}.  SSL can achieve better performance than SL trained models with much less labels \cite{chen2020simple}. There are four main types of SSL methods, innate relationships, generative, contrastive and self-prediction. Innate relationships leverage the internal structure of the data, e.g. predicting the rotation angle, or finding the position of shuffled image patches. Generative methods use autoencoders or generative adversarial networks (GANs). They are assessed by the quality of reconstructing the original images. Contrastive methods rely on using data augmentation to create positive pairs of an image. Different images are treated as negative pairs. Various distance metrics have been proposed to keep positive pairs close and negative pairs distant in the embedding space. Some examples include SimCLR \cite{chen2020simple}, MoCo \cite{he2020momentum}, BYOL \cite{grill2020bootstrap}, SimSiam \cite{chen2021exploring}, DINO \cite{caron2021emerging}. Self-prediction methods mask out portions of the original image and try to reconstruct the original image. The main difference between self-prediction and generative methods is that self-prediction applies augmentation on portions of the original image while generative methods apply to the whole image. SSL application in medical imaging is rapidly rising, as medical labels are costly and time-consuming to obtain. SSL can speed up the process of model development.

\subsection{Unsupervised domain adaptation}
UDA methods can be divided into two types, moment matching and adversarial training. Moment matching methods try to match the first or second moments between the hidden activation distributions of the source and target domains \cite{zellinger2017central}. Some examples are deep adaptation networks (DAN) \cite{long2015learning}, joint adaptation networks (JAN) \cite{long2017deep}, correlation alignment \cite{sun2016deep}, and central moment discrepancy (CMD) \cite{zellinger2017central}. Adversarial training methods try to learn domain invariant features. It trains a domain classifier to put the source and target domain features in the same space. The most used adversarial training method is domain adversarial neural networks (DANN) \cite{ganin2015unsupervised}. Other methods build upon it such as adversarial discriminative domain adaptation (ADDA) \cite{tzeng2017adversarial}, maximum classifier discrepancy (MCD) \cite{saito2018maximum}, batch spectral penalization (BSP) \cite{chen2019transferability}, and minimum class confusion (MCC) \cite{jin2020minimum}.           

There are works of applying UDA in medical image analysis focusing on classification \cite{ahn2020unsupervised}, object localization, and segmentation \cite{kamnitsas2017unsupervised}. More specifically for skin lesion classification, Chamarthi et al. compare performance of different UDA methods \cite{chamarthi2024mitigating}. Wang et al. demonstrate applying UDA methods can mitigate model bias against minority groups \cite{wang2024achieving}. 

\subsection{Active domain adaptation}
UDA methods often exhibit lower performance compared to their supervised counterparts \cite{tsai2018learning, chen2018domain}. Supervised domain adaptation (SDA) requires labels of the target domain. However, for the medical imaging field especially, obtaining large amounts of labels can be cost intensive. Therefore, selecting the most informative samples for annotation is crucial. This is where active learning (AL) comes in. AL is trying to maximize a model's performance given a limited amount of labels.  Active domain adaption (ADA) combines active learning with domain adaptation. Different active learning methods have different sampling strategies. Broadly, they are based on the two types of metrics, uncertainty \cite{lewis1994heterogeneous, scheffer2001active}, and diversity \cite{dutt2016active, hoi2009semisupervised}. Uncertainty-based methods choose instances with the high uncertainty based on measures such as entropy, classification margins, or confidence \cite{ducoffe2018adversarial, gal2017deep, wang2014new}. Diversity-based methods choose instances that are representative of the entire dataset. They optimize for diversity in the embedding space with clustering or core-set selection \cite{sener2018active, geifman2017deep, gissin2019discriminative, sinha2019variational}. Several methods attempt to combine these two types of metrics \cite{prabhu2021active, huang2023divide, fu2021transferable, ash2020deep}. Zhao et al. applied SSL and AL for skin lesion segmentation, but there is no domain adaptation \cite{zhao2022self}. No prior work exists on applying ADA methods to skin lesion classification.


\section{Methodology}
\begin{figure}[htbp]
  \centering 
  \includegraphics[width=0.8\linewidth]{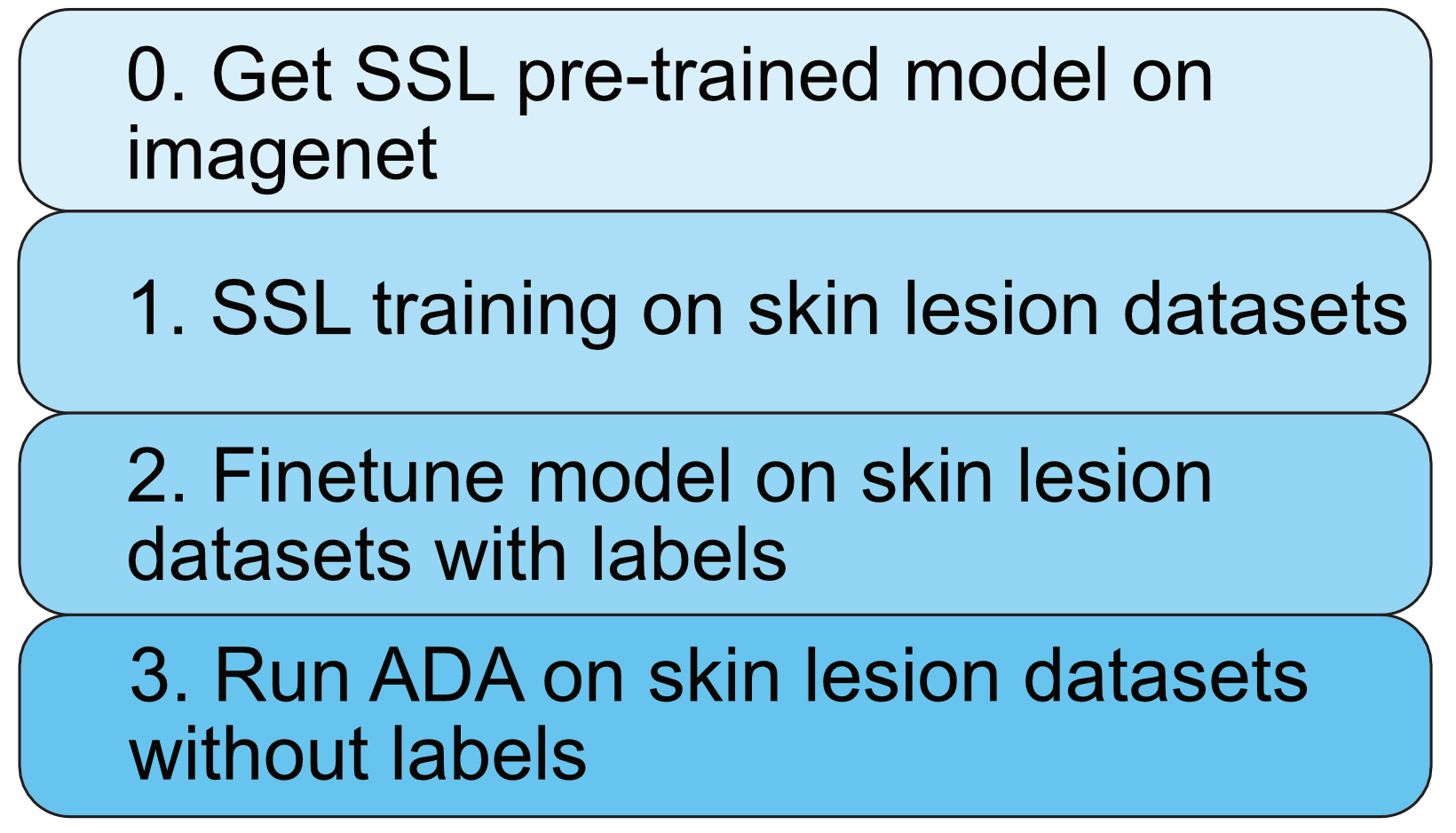}
  \caption{Proposed workflow.}
  \label{fig:method_steps}
\end{figure}

Here, we propose a workflow combining SSL and DA methods to improve the generalization of skin lesion classification models. Figure \ref{fig:method_steps} presents the steps of the workflow. 

\begin{figure}[htbp]
  \centering 
  \includegraphics[width=1.\linewidth]{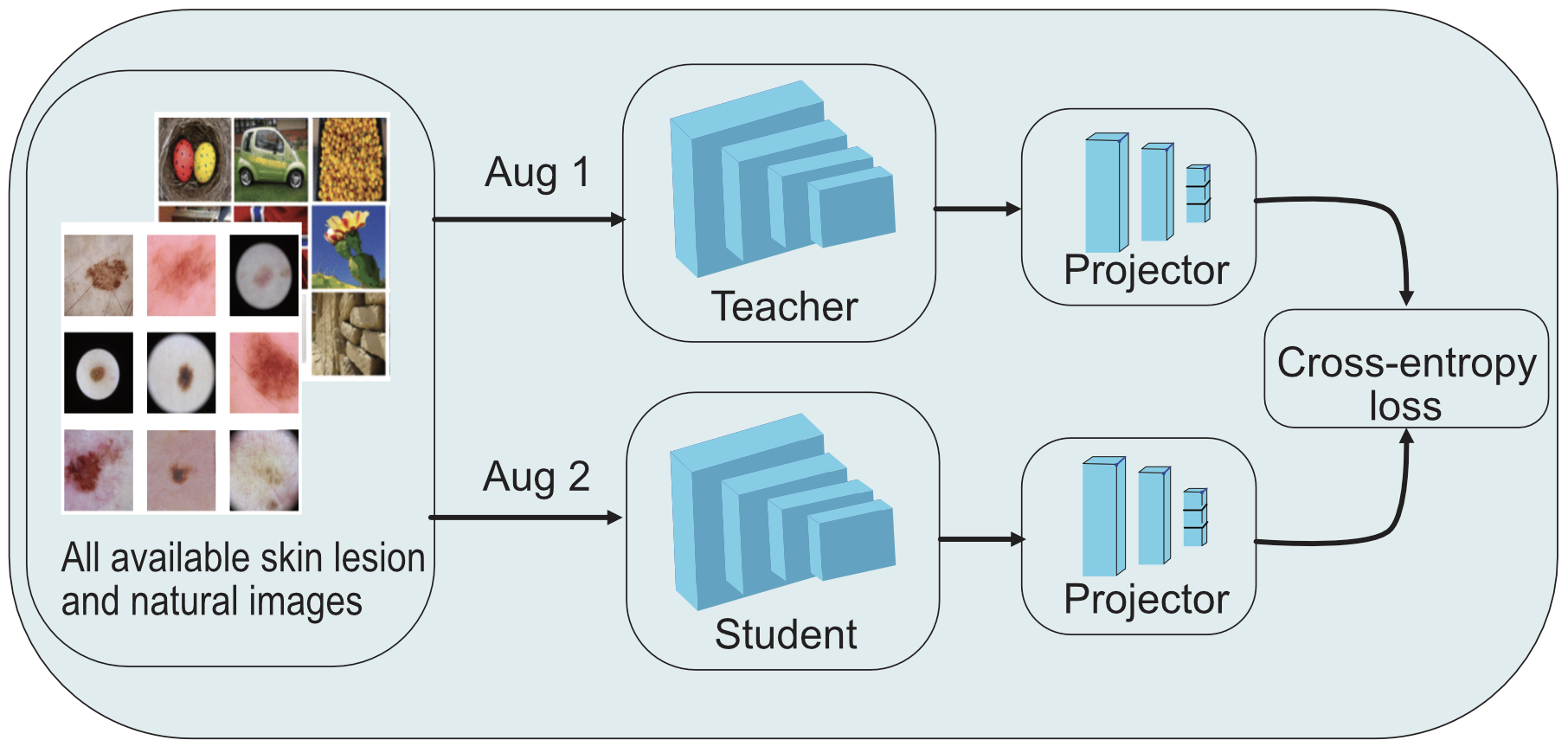}
  \caption{DINO SSL architecture.}
  \label{fig:dino_architecture}
\end{figure}

The initial step involves selecting an SSL pre-trained model on natural image datasets, such as ImageNet. Here we select the DINO SSL method since it’s shown to produce the SOTA performance on ImageNet among many SSL methods, e.g. SimCLR, MoCov2, Barlow Twins, BYOL, SwAV \cite{oquab2023dinov2, caron2021emerging}. Figure \ref{fig:dino_architecture}  shows the DINO architecture. It uses knowledge distillation with a student and teacher architecture. Aug 1 and Aug 2 represent different augmented views of an image, which are fed into the student and teacher networks, respectively. Both networks have the same model architecture. The loss term minimizes the cross-entropy between the features learned by the student and teacher networks. Centering and sharpening are applied on the teacher’s output to prevent the model from learning trivial solutions. It doesn’t require a large batch size or negative samples. We choose ResNet50 \cite{he2016deep} as the backbone model because it’s used in other similar studies \cite{chamarthi2024mitigating}, allowing for direct comparison of results. The projector head has a couple linear layers with the hidden dimension as 256 and the output dimension as 65536.   
The DINO cross-entropy loss term between the teacher’s and student’s learned features is defined as follows:
$$minH(P_t(x), P_s(x)) = \min{-(P_t(x)\log(P_s(x)))}$$

$P_t(x), P_s(x)$ are the probability distribution from the output of the teacher and student network respectively.

The first step involves continuing DINO training on skin lesion datasets. Because no labels are required, training can be performed on all the available skin lesion data. This facilitates the model’s generalization from natural images to skin lesion data. 

The second step involves finetuning the model on a labeled skin lesion dataset. Here, we focus on a common task: binary classification of melanoma (cancer) versus nevus (benign). We freeze the backbone model and fine-tune only the linear classification layer to isolate the impact of feature quality, excluding the influence of a more complex classifier. 
    
The third step involves applying ADA. An optional preliminary step involves applying UDA to align the source and target domains without target domain labels. One popular UDA method is DANN \cite{ganin2015unsupervised}, which has shown good performance among UDA methods applied to skin lesion datasets \cite{chamarthi2024mitigating}. DANN is an adversarial training based method. It comprises three main components: a feature extractor, a class classifier, and a domain classifier. A gradient reversal layer between the domain classifier and the feature extractor encourages the feature extractor to learn domain invariant features.   


\begin{figure}[htbp]
  \centering 
  \includegraphics[width=1.\linewidth]{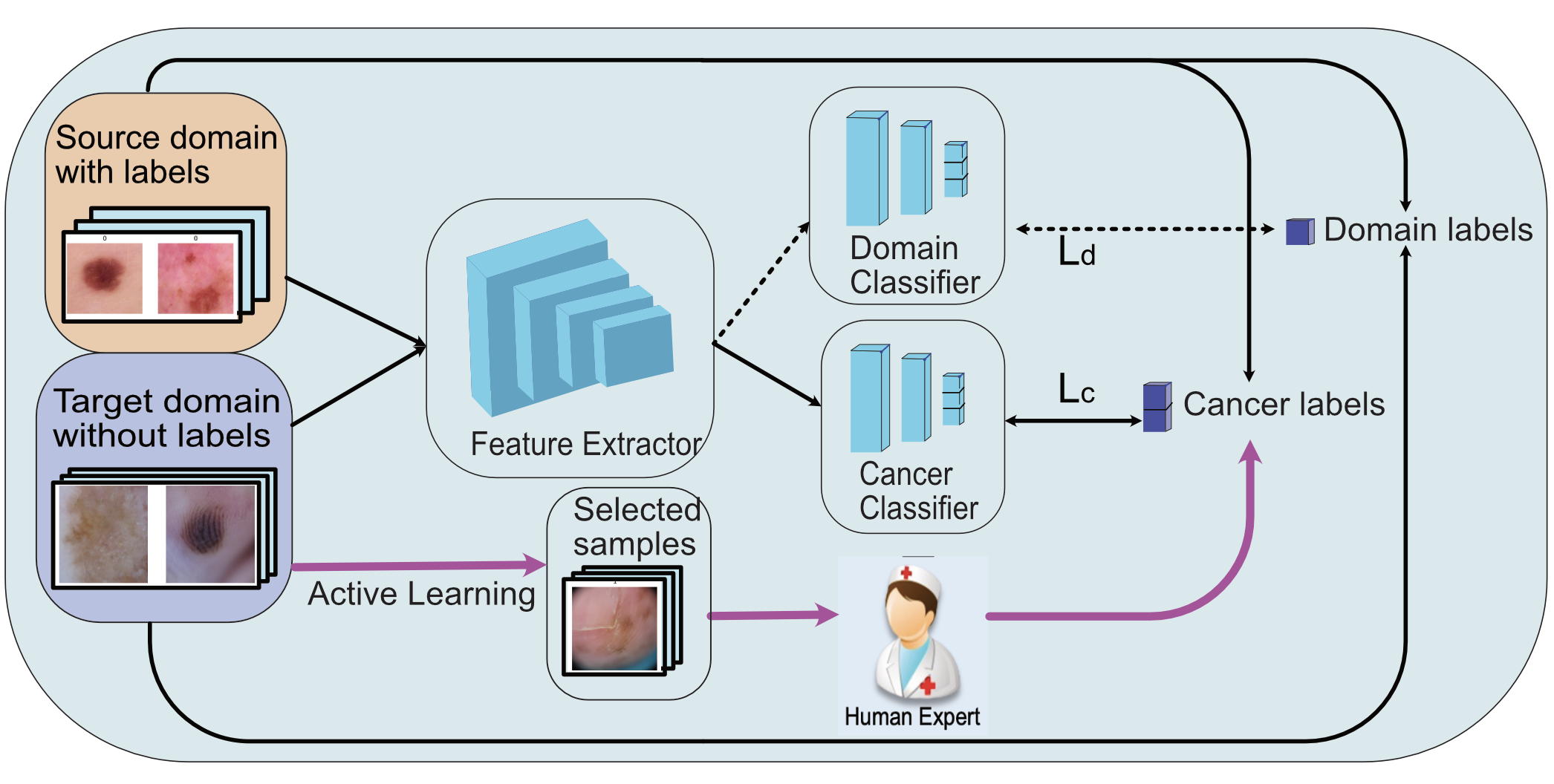}
  \caption{Active domain adaptation architecture.}
  \label{fig:ADA_architecture}
\end{figure}

We compare several ADA methods: AADA \cite{su2020active}, CLUE \cite{prabhu2021active}, BADGE \cite{ash2020deep}, and random sampling. Figure \ref{fig:ADA_architecture} illustrates the general ADA architecture. We have skin lesion labels for the source domain dataset, but lack labels for the target domain datasets. Both source domain and target domain datasets can be input to the model, resulting in two computed losses. $L_d$ represents the domain loss, which we use to train the model to be unable to distinguish the domains. $L_c$ represents the skin lesion classification loss. Cross-entropy is used to calculate both losses. We apply active learning to select informative samples from the target domain, which are then sent to human experts for annotation. Using these annotated labels, we can then train the model using the classification loss. The domain classifier is optional; adversarial training-based ADA methods, such as AADA, utilize it, while other methods don't. 

For example, the active learning sampling criterion of AADA method is as such:
$$S(x) =\frac{1-G_d(G_f(x))}{G_d(G_f(x))}H(G_y(G_f(x)))  $$
It incorporates both the diversity cue $\frac{1-G_d(G_f(x))}{G_d(G_f(x))}$, and uncertainty cue $H(G_y(G_f(x)))$. 
 

\section{Experiment}
To demonstrate the effectiveness of our approach in improving model generalization, we conduct tests on a set of skin lesion datasets exhibiting domain shifts. We compare our results with the existing approach using UDA methods. We demonstrate the benefits of SSL training and applying ADA respectively. When applying the ADA method, we consider real-world clinical usage by employing a small sampling size of 10. Given an additional labeling budget, we can employ an iterative training strategy with an increased number of provided samples.

\subsection{Dataset}
Fogelberg et al. \cite{fogelberg2023domain} created skin lesion datasets representing different domains by grouping data from HAM10000 \cite{tschandl2018ham10000}, BCN20000 \cite{hernandez2024bcn20000}, and MSK \cite{cassidy2022analysis} according to biological traits such as patient age and lesion location. They quantified the domain shift intensity between groups using cosine similarity and Jensen-Shannon divergence. The details of the grouped datasets are presented in Table \ref{tab:biological_factors}. It lists a total of 11 datasets. Dataset H is the source domain, and the other 10 datasets are target domains. It also shows the number of samples in each dataset, the ratio of melanoma and nevus, and biological traits.

\begin{table*}[h]
    \caption{Skin lesion datasets of different domains. }
    \centering
    \begin{tabular}{|c|c|c|c|c|c|}
        \hline
        \textbf{Abbreviation} & \textbf{Origin} & \textbf{Biological factors} & \textbf{Melanoma amount} & \textbf{Nevus amount} & \textbf{Total sample size} \\ \hline
        H & HAM & $Age <= 30$, Loc. = Body (default) & 465 (10\%) & 4234 (90\%) & 4699 \\ \hline
        HA & HAM & $Age > 30$, Loc. = Body & 25 (4\%) & 532 (96\%) & 557 \\ \hline
        HLH & HAM & $Age < 30$, Loc. = Head/Neck & 90 (45\%) & 121 (55\%) & 220 \\ \hline
        HLP & HAM & $Age > 30$, Loc. = Palms/Soles & 15 (7\%) & 203 (93\%) & 218 \\ \hline
        B & BCN & $Age > 30$, Loc. = Body (default) & 1918 (41\%) & 2721 (59\%) & 4639 \\ \hline
        BA & BCN & $Age <=30$, Loc. = Body & 71 (8\%) & 808 (92\%) & 879 \\ \hline
        BLH & BCN & $Age > 30$, Loc. = Head/Neck & 612 (66\%) & 320 (34\%) & 932 \\ \hline
        BLP & BCN & $Age > 30$, Loc. = Palms/Soles & 192 (65\%) & 105 (35\%) & 297 \\ \hline
        M & MSK & $Age > 30$, Loc. = Body (default) & 565 (31\%) & 1282 (69\%) & 1847 \\ \hline
        MA & MSK & $Age <=30$, Loc. = Body & 37 (8\%) & 427 (92\%) & 464 \\ \hline
        MLH & MSK & $Age > 30$, Loc. = Head/Neck & 175 (60\%) & 117 (40\%) & 292 \\ \hline
    \end{tabular}
    \label{tab:biological_factors}
\end{table*}

\begin{table*}[h]
    \caption{Comparison of our SSL methods together with the BSP UDA method and baseline reported from Chamarthi et al. \cite{chamarthi2024mitigating}.}
    \centering
    \begin{tabularx}{\textwidth}{p{1.1cm}cccccccccc}
    \toprule
    {} & HA & HLH & HLP & B & BA & BLH & BLP & M & MA & MLH  \\
    \midrule
    Mel (ratio) & 0.04 & 0.45 & 0.07 & 0.41 & 0.08 & 0.66 & 0.65 & 0.31 & 0.08 & 0.6 \\
    \midrule
    Baseline & 0.14$_{\pm0.02}$ & 0.69$_{\pm0.04}$ & 0.37$_{\pm0.15}$ &   0.57$_{\pm0.02}$ & 0.19$_{\pm0.06}$ & 0.73$_{\pm0.03}$ & 0.77$_{\pm0.05}$ & 0.34$_{\pm0.01}$ & 0.15$_{\pm0.04}$ & 0.68$_{\pm0.03}$ \\
    BSP & 0.16$_{\pm0.03}$ &  0.82$_{\pm0.02}$ &  0.65$_{\pm0.04}$ & 0.75$_{\pm0.01}$ & \textbf{0.34$_{\pm0.05}$} & \textbf{0.86$_{\pm0.01}$} & 0.83$_{\pm0.02}$ & 0.46$_{\pm0.02}$ & 0.17$_{\pm0.03}$ & 0.73$_{\pm0.01}$  \\
    \midrule
    Dino retrained & \textbf{0.17$_{\pm0.04}$} & 0.86$_{\pm0.01}$ &  0.5$_{\pm0.09}$ & \textbf{0.77$_{\pm0.01}$} & 0.31$_{\pm0.09}$ & 0.84$_{\pm0.01}$ & \textbf{0.86$_{\pm0.02}$} & \textbf{0.58$_{\pm0.02}$} & \textbf{0.24$_{\pm0.05}$} & 0.75$_{\pm0.04}$  \\
    Dino pre-trained & 0.13$_{\pm0.04}$ & \textbf{0.89$_{\pm0.03}$} &  0.6$_{\pm0.15}$ & 0.76$_{\pm0.01}$ & \textbf{0.34$_{\pm0.07}$} & 0.83$_{\pm0.01}$ & 0.83$_{\pm0.02}$ & 0.52$_{\pm0.01}$ & 0.12$_{\pm0.02}$ & \textbf{0.77$_{\pm0.01}$}  \\
    SL pre-trained & 0.16$_{\pm0.04}$ & 0.79$_{\pm0.03}$ &  \textbf{0.71$_{\pm0.12}$} & 0.73$_{\pm0.01}$ & 0.33$_{\pm0.05}$ & 0.83$_{\pm0.02}$ & 0.80$_{\pm0.02}$ & 0.52$_{\pm0.01}$ & 0.20$_{\pm0.02}$ & 0.70$_{\pm0.02}$  \\
    \bottomrule
    \end{tabularx}
    \label{tab:ssl_sl_comp}
\end{table*}

\subsection{Metrics}
We use AUPRC metric because it effectively handels imbalanced datasets given greater weight to the cancer class. As shown in Table \ref{tab:biological_factors}, many of the skin lesion datasets are highly imbalanced, with the cancer class comprising as little as 4\% of the data. We don’t use accuracy or AUROC because they are only suitable for balanced datasets. However, we must consider the varying AUPRC baselines across datasets, as these baselines reflect the different melanoma ratios.

\subsection{Model Training}
We obtained the pre-trained DINO Resnet50 model from Facebook Research’s GitHub repository \footnote{https://github.com/facebookresearch/dino?tab=readme-ov-file}. When retraining the model with DINO SSL on the skin lesion data, we utilized all datasets listed in Table \ref{tab:biological_factors}. We employed the default DINO data augmentations, with random cropping, two larger view sizes (0.4-1.0 scale) and four smaller view sizes (0.05-0.4 scale), rotation, jitter, blur and black and white transformations. Only the larger view sizes are passed to the teacher model, whereas all the view sizes are passed to the student model. This encourages the model to learn “local-to-global” correspondence. We explored various data augmentation strategies, and found that the default setting yielded the best results. Our optimized training process involves initially freezing the backbone model and training only the prediction head layers with a starting learning rate (LR) of 1e-3. We trained for 30 epochs. Subsequently, we trained all the layers using adaptive learning rates, with initial LR values of 1e-4 for the backbone layers and 5e-4 for the projector layers. We employed the one-cycle LR scheduler. We trained for 25 epochs at which point the loss plateaued.

We then fine-tuned the model on the skin lesion source domain dataset, H. We froze the backbone and trained only the classifier head (a single linear layer), to isolate the impact of the backbone feature quality.       

Model training was performed on a RTX 4090 GPU. The batch size was 128, and the image size was 224 x 224 pixels.

\section{Results and discussions}

We compared the skine lesion classification performance of different methods on target domains using the AUPRC metric. We present the comparison results alongside the baseline (no domain adaptation) and the BSP UDA method reported in \cite{chamarthi2024mitigating}. Table \ref{tab:ssl_sl_comp} presents the results, with the mean and standard deviation reported across 5 random seeds.

First, compared with the results reported in \cite{chamarthi2024mitigating} for the baseline and BSP methods, our approach achieved superior performance on 9 of the 10 target domains, with slightly worse results observed only for BLH dataset. This demonstrates that SSL retraining on in-domain data is an effective UDA method. 

Next, we compared the performance of the SL pre-trained model with the SSL Dino pre-trained model on ImageNet. The SSL Dino pre-trained model could be further retrained on the skin lesion datasets. We observed that the SL pre-trained model performed worse than the SSL pre-trained model on 9 of the 10 datasets. Therefore, SSL pre-training is superior to SL pre-training for domain adaptation. Furthermore, the SSL Dino retrained model exhibited improved performance on 6 of the 10 datasets compared to the model without retraining. Thus, retraining on in-domain data is advantageous for domain adaptation.

\begin{figure}[htbp]
  \centering 
  \includegraphics[width=1.\linewidth]{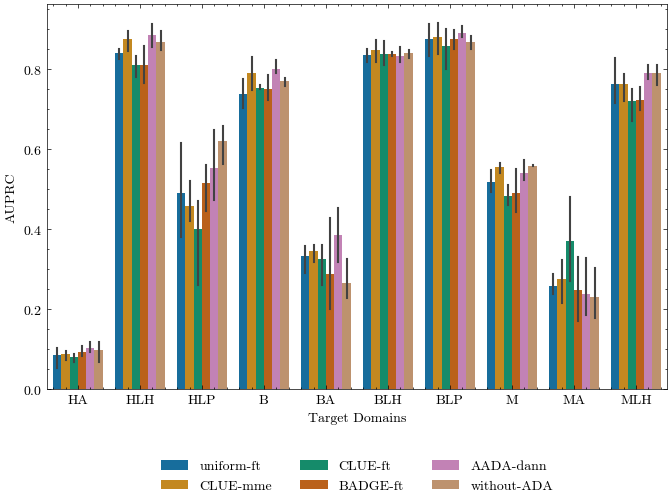}
  \caption{Comparing ADA methods with AUPRC value.}
  \label{fig:ada_auprc}
\end{figure}

\begin{figure}[htbp]
  \centering 
  \includegraphics[width=1.\linewidth]{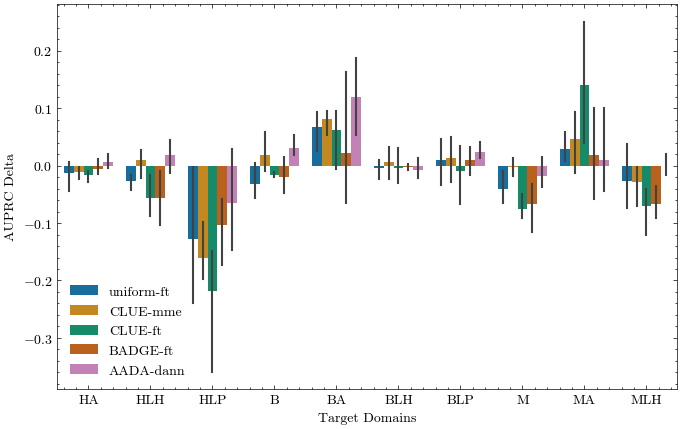}
  \caption{Comparing ADA methods.}
  \label{fig:ada_delta}
\end{figure}

Here, we present the results of applying ADA methods with 10 annotated samples from the target domain. Figure \ref{fig:ada_auprc} shows the AUPRC achieved by the ADA methods compared to the baseline (without ADA), which corresponds to the DINO retrained model from Table \ref{tab:ssl_sl_comp}. 

We compared five different ADA methods. The ADA methods comprise two components: active learning and domain adaptation. The active learning methods we evaluated included CLUE, AADA, BADGE and uniform sampling. The domain adaptation methods we evaluated included MME \cite{saito2019semi}, DANN, and fine-tuning. The simplest ADA method involves uniform (random) sampling and fine-tuning. 

Significant performance variability exists across the different target domains, primarily due to the varying AUPRC baselines, which are directly related to the melanoma ratios. We observed performance variability for the same ADA methods across the target domains. This variability primarily stems from the domain shift relative to the source domain.   

To better visualize the difference of the five ADA methods, Figure \ref{fig:ada_delta} presents the AUPRC delta between the ADA methods and the baseline. The AADA-DANN method exhibited the best performance. It performed better than or similar to the baseline on 9 of the 10 datasets, with the exception of the HLP dataset. One potential explanation is the small number of melanoma samples (only 15) in the HLP dataset, leading to greater variance in performance.

\section{Conclusion}
In this paper, we proposed a workflow to improve the generalization of skin lesion classification models by combining SSL and ADA methods. Our results demonstrated the benefits of each method. SSL proved to be an effective UDA method, and ADA provided further performance improvements. We compared five ADA methods and showed that AADA-DANN yielded the best performance. We evaluated the performance on 10 different datasets exhibiting varying degrees of domain shift. This mimics real-world clinical scenarios, where domain shifts are common. Thus, our approach can facilitate wider clinical adoption of the skin lesion classification models. Future studies could explore different backbone model architectures, such as vision transformers, given their demonstrated strong performance on natural image datasets.






{\small
\bibliographystyle{ieee_style}
\bibliography{root}
}

\end{document}